# Predicting Business Angel Early-Stage Decision Making Using AI

Yan Katcharovski, Andrew L. Maxwell

York University, Toronto, Canada

**Abstract**

External funding is crucial for early-stage ventures, particularly technology startups that require significant R&D investment. Business angels offer a critical source of funding, but their decision-making is often subjective and resource-intensive for both investor and entrepreneur. Much research has investigated this investment process to find the critical factors angels consider. One such tool, the critical factor assessment (CFA), deployed more than 20,000 times by the Canadian Innovation Centre, has been evaluated post-decision, and found to be significantly more accurate than investors' own decisions. However, a single CFA analysis requires three trained individuals and several days, limiting its adoption. This study builds on previous work validating the CFA to investigate whether the constraints inhibiting its adoption can be overcome using a trained AI model. In this research, we prompted multiple large language models (LLMs) to assign the eight CFA factors to a dataset of 600 transcribed, unstructured, startup pitches seeking BA funding with known investment outcomes. We then trained and evaluated machine learning classification models using the LLM-generated CFA scores as input features. Our best-performing model demonstrated a very high predictive accuracy (85.0% for predicting BA deal/no-deal outcomes) and exhibited very significant correlation (0.896 Spearman's ρ, $p < .001$) to conventional human-graded evaluations. The integration of AI-based feature extraction with a structured and validated decision-making framework yielded a scalable, reliable, and less-biased model for evaluating startup pitches, removing the constraints that limited its adoption earlier.

**Keywords** Business angels, Entrepreneurial finance, Critical Factor Assessment (CFA), Large Language Models (LLMs), Artificial Intelligence, Startup Pitch Evaluation, Investment Decision-Making

## 1. Introduction

Acquiring external funding is a critical milestone for high-potential ventures, allowing entrepreneurs to convert innovative concepts into scalable enterprises while funding the development and launch of their business prior to revenue generation [1]. Business angels (BAs) are a pivotal source of early-stage financing, acting as private investors who contribute personal capital for equity. BAs frequently serve as the initial external equity source, facilitating the early-stage development and validation of the business through seed funding that can subsequently be enhanced through venture capital (VC) investment [2]. Their contributions go beyond financial assistance to providing mentorship, strategic counsel, and access to valuable networks that profoundly impact a venture's growth trajectory [3]. BAs are often motivated by the desire to give back to the community and provide support to the next generation of entrepreneurs. However, the process of obtaining BA funding is characterized by significant complexity. Entrepreneurs, especially first-time founders, face a decision-making environment influenced by subjective assessments, insufficient historical data, and significant information asymmetries, often complicated by conflicting sources of guidance [4].

Research has identified a large number of factors that can affect BA investment decisions. Misunderstanding arises due to a lack of awareness of the decision process, especially how the impact of different factors (such as market potential, entrepreneurial experience, product viability, and financial projections) change as the process evolves [5, 6]. Even in the case of an agreement on individual factors, the assessment of these factors is often based on the investor's individual experience. BAs use heuristics influenced by personal experiences and biases, which can result in inconsistencies in their assessment of the likely success of a venture or of the funding decision [7]. Further, the lack of rigor (or a framework) in the process limits the potential for valuable, objective, and consistent feedback. The validated critical factor assessment (CFA) evaluation framework is reliable and can be used to provide specific and relevant guidance to the entrepreneur. However, its use in practice has been hampered by the time-consuming and resource-intensive characteristics of its deployment.

This study builds on previous research on the CFA framework, widely used and validated in entrepreneur development and investment assessments, to develop an AI operationalization model. We addressed the challenges of widespread deployment and scalability of the original tool, given the previously identified implementation constraints of cost, training, and potential biases [6, 8]. The results of this research reinforce and replicate the original deployment of the CFA tool, using the same eight factors as in Maxwell's original research: Features & Benefits, Readiness, Market Size, Barriers to Entry, Supply Chain, Entrepreneurial Experience, Financial Expectations, and Adoption [6]. This study provides evidence of how well-designed and trained AI tools can provide valuable insights for entrepreneurs and those working with them to identify specific venture challenges that must be addressed to increase the likelihood of receiving funding.

This study was facilitated by our access to the original data and assessment techniques used in prior work [6], knowledge of the CFA process, advancements in large language models (LLMs), and supervised machine learning algorithms. This allowed us to automate the evaluation of startup pitches, synthesize factors that impact BA decisions, and provide diagnostic feedback to the entrepreneur. The dataset included transcripts from 600 pitches presented on the U.S. television show *Shark Tank*, providing a diverse sample of real-world interactions between investors and entrepreneurs, resembling those in conventional funding settings. The validity of this dataset to replicate real decision-making was discussed in Maxwell's research that studied full, behind-the-scenes pitches from the Canadian television show *Dragon's Den* (briefly addressed in this paper) [6].

Our research demonstrates that an AI-augmented CFA evaluation tool, using Maxwell's eight-factor CFA framework, offers superior reliability in forecasting investment decisions, surpassing conventional models in both speed and objectivity while preserving robust predictive accuracy. It reinforces many of the insights from Maxwell's original research, replicating its accuracy (although without the detailed evolution of the full BA investment-decision process). Drawing on the research that led to the creation of the CFA, this study demonstrates how AI can help prepare funds-seeking entrepreneurs for interactions with angels, encouraging them to consider each of the eight critical factors. It highlights the value of understanding why ventures get rejected, contesting the presuppositions about the constraints of algorithmic assessment in scenarios requiring nuanced judgment by bolstering the ability to offer unbiased feedback, and reinforcing Noble laureate Daniel Kahneman's original insights when referring to the CFA as a mechanism designed to facilitate decision making [7, p. 216].

The study demonstrates a novel approach, by combining a validated venture evaluation framework with an LLM and showing that a framework-aware AI outperforms a generalist AI, thus validating the original research that facilitated the development of the CFA (a framework is more accurate than no framework). Our results offer a number of practical applications and future research implications, for entrepreneurs and investors, and their stakeholders keen to rapidly iterate their ventures to increase their likelihood of venture success (and funding), as well as to those considering how AI can enhance existing entrepreneurship insights and tools to guide those interested in pursuing entrepreneurial careers. This offers a promising area for future research — the ability to leverage existing research to provide AI powered tools to coach or mentor entrepreneurs through their venture creation journey "on demand" and at relatively low cost. Ultimately, this not only allows us to leverage validated research into entrepreneurship education but also increases the likelihood of successful venture development and the ability to attract BA investment where required.

## 2. Why Business Angel Funding is Important

A BA is a wealthy individual who typically invests their own money in exchange for equity in early-stage startups. In 2023, the average BA investment size was $339,390 in return for 9.7% of startup equity [21]. In contrast to initial funding from family and friends, which is typically influenced by personal relationships, BA investments are based on evaluations of a venture's growth potential and financial return prospects as well as the psychological motivation they experience from helping next-gen entrepreneurs [17]. Unlike venture capitalists (VCs), BAs do not have to justify their decisions to anyone; instead, they allocate their personal capital in return for equity interests, making strategic

decisions based on considerations such as market opportunity, competitive advantage, and competencies of the entrepreneurial team [6, 14–16].

BA funding plays a crucial role in the entrepreneurial ecosystem, especially during the initial phases of venture development when external capital is essential for survival and expansion [2,9]. This is particularly the case for high-growth enterprises, often linked to the development of technologically innovative solutions (and disruptive business models), which can incur significant costs associated with product development, market entry, and operational scaling—expenditures that generally occur prior to sufficient revenue generation [10–12]. BAs play a crucial role in bridging this funding gap, referred to as the "valley of death", by supplying the capital required to connect the initial concept development phase to commercial viability [13].

BA participation encompasses more than financial contribution, as their investment frequently acts as significant validation signals for other stakeholders, including customers and partners. VCs, who come in later, often regard BA support and their relationship with the venture as a prerequisite for subsequent funding, as a signal of the future of their own potential relationship. Thus, BA funding serves as an essential conduit, facilitating the transition of startups from the seed stage to the acquisition of institutional capital [5, 17–19]. Given that BA investors typically possess significant experience as entrepreneurs or industry veterans, they are seen as "smart money" helping the entrepreneur in avoiding pitfalls and establishing industry connections [20]. In a way, BA investment illustrates an experimental approach to reducing uncertainty, as they finance early-stage and untested ventures in anticipation of future investment opportunities [2, 5].

In 2023, U.S. angel investors invested $18.6 billion in 54,735 ventures [21], representing 7% of the total venture financing. Although this amount is significantly lower than the $248.4 billion invested by VC firms across 29,303 deals during the same timeframe [22], BAs participated in the majority of early-stage investment transactions, accounting for 65% of all seed-stage funding deals. Importantly, many BA investments are made in anticipation of raising future funds from VCs or elsewhere—BAs tend to consider some of the critical factors that would influence future investment decisions as well and relate to actual business performance [19, 20].

However, the BA investment process, despite its importance, is fraught with deficiencies such as subjective decision-making, information asymmetries, and inconsistent evaluation criteria, leading to suboptimal funding outcomes, as discussed next [6, 7, 23].

## 3. Previous Research on BA Decision-Making

Early research on BA decision-making has concentrated on determining the primary criteria that affect investment results. Factors such as market potential, financial projections, entrepreneurial experience, product viability, and the strength of the business model are critical determinants of funding decisions [6, 24–25]. Much of the foundational research has depended on retrospective self-reports from investors, which present challenges such as hindsight and confirmation bias, likely to distort perceptions of prior decisions [26]. Research in behavioral decision-making, popularized by Kahneman in his book *Thinking Fast and Slow*, indicates that heuristics such as anchoring, availability bias, and overconfidence, frequently result in suboptimal judgments in environments characterized by high uncertainty, exemplified by early-stage investing [7]. Observational studies of investor–entrepreneur interactions have been suggested as a method to reduce biases in self-reported data [27]. The transition to real-time analysis has guided the creation of structured evaluation frameworks designed to improve objectivity in investment decisions.

The critical factor assessment (CFA) framework, developed by the Canadian Innovation Centre (CIC) in the 1980s, represents a significant contribution in the domain of evaluating the commercial viability of new ventures. The initial framework comprised 37 evaluation criteria including market need, technical feasibility, and management capability [8], and was refined by Dr. Andrew Maxwell to eight critical factors: Product Adoption, Product Status, Protectability, Customer Engagement, Route to Market, Market Potential, Relevant Experience, and Financial Model. This approach aims to enhance the predictive accuracy of investment decisions and decrease cognitive load on evaluators [6].

Between 1976 and 2004, the Canadian Innovation Assistance Program (IAP), in collaboration with the CIC, used the CFA to evaluate over 14,000 innovative venture applications, proving its robustness and adoption [28].

Empirical studies on Maxwell's CFA framework (hereafter, the CFA) have shown its efficacy in structuring BA decision-making. An investor behavior analysis in quasi-experimental settings, exemplified by the Canadian reality television show *Dragons' Den*, indicated that BAs utilize elimination-by-aspects (EBA) heuristics, leading to the immediate rejection of opportunities with critical flaws [6]. This heuristic screening process allows investors to effectively narrow down the ventures for investment, thus accelerating decision-making. The CFA framework yielded a predictive accuracy of 87.5% when forecasting which ventures progress beyond the initial selection stage, as BAs consistently rejected pitches that score poorly on critical factors [6]. The CFA was also found to be an accurate predictor of whether early-stage ventures would go on to succeed at go-to-market product launches. A study that followed 561 CFA-screened ventures five years post-assessment found that its critical factors predicted venture success with an accuracy of 80.9% [29].

While the CFA framework has significantly influenced BA decision-making, subsequent research has explored its applications and limitations. Structured evaluation frameworks, such as those modeled on the CFA, enhance decision-making efficiency in BA syndicates during due diligence [30]. Concerns persist that the strict application of these frameworks may unintentionally reinforce biases or overlook the nuanced potential of atypical, and often successful, ventures [31]. A systematic review of gender disparities in entrepreneurial equity financing indicated that biases, both conscious and unconscious, can affect BA decisions, especially during early-stage evaluations [31].

Structured tools such as the CFA help reduce biases by offering objective benchmarks; however, the effectiveness of these frameworks is contingent upon the degree to which their foundational criteria are devoid of inherent bias. Should the critical factors embody implicit biases, structured decision tools may unintentionally sustain, rather than address, inequities in early-stage funding decisions.

## 4. Limitations of CFA Deployment and Potential of AI

The CFA framework effectively provides structured evaluations of early-stage ventures; however, several inherent constraints limit its scalability, efficiency, and broader applicability. The limitations arise from the framework's reliance on manual human assessment [6] and dependence on data quality, resulting in challenges especially evident in the fast-paced, high-volume contemporary entrepreneurial finance [32]. The process is time-consuming and resource-intensive, with expert-driven CFA assessments typically requiring up to six weeks and costing up to $1,400 per evaluation, creating substantial obstacles for resource-limited startups and organizations with restricted evaluation capabilities [33]. Entrepreneurs must also make decisions rapidly as markets change and opportunities arise, with studies indicating that moving quickly is a winning strategy in innovative early-stage startups. For such a startup to succeed, rapid iterations and decision making (often referred to as a Lean Startup approach, coined by author Eric Ries) must be paired with the ability to focus on the right issues, while ignoring the wrong ones [34]. Early-stage startups do not possess sufficient data to support their business assertions, especially on market validation, financial forecasts, and operational preparedness [35, 36]. This problem is exacerbated by information asymmetry, when entrepreneurs may inadvertently exclude essential details or, in certain instances, deliberately withhold information to portray their ventures more favorably [37, 38].

A second significant limitation is the subjective nature of human assessments. Cognitive biases, including anchoring, overconfidence, and availability heuristics, can still impact investment decisions, despite the structured criteria inherent in the CFA [39–41]. Kahneman emphasized that even skilled evaluators are vulnerable to biases that can impair judgment, especially in high-uncertainty situations that demand swift decision-making [7]. This challenge is intensified in competitive entrepreneurial ecosystems, where swift evaluations are crucial for seizing time-sensitive opportunities and sustaining deal flow momentum [42].

The status quo of the market offers access to a validated, useful and relatively easy-to-use assessment tool that was expected to be useful for thousands of entrepreneurs and entrepreneurship centers, but the constraints (and costs) of deployment made this impractical. Advancements in AI offer opportunities to address this constraint.

Research interest in AI's potential capabilities in investment decision-making has grown recently. An expertly developed and deployed AI model can efficiently process large datasets with speed and consistency, while minimizing the biases that frequently influence human judgment [43, 44]. These models are further capable of assessing an extensive range of inputs, including quantitative financial metrics and the nuances of qualitative pitch content, at speeds and scales surpassing manual methods [45]. In contrast to static decision frameworks, AI models possess the capability to continuously revise their evaluations based on new data, changing market conditions, and emerging trends [46]. Adaptability is essential to the new venture ecosystem characterized by uncertainty, rapid technological changes, and evolving consumer behaviors [47, 48]. As AI systems analyze larger datasets, their predictive models offer investors progressively precise insights derived from new data [46].

Some research exists on the use of machine learning (ML) techniques to predict investor decision-making, but the lack of a rigorous framework (such as the CFA) for analysis limits accuracy. Artificial neural networks (ANNs) and random forests techniques have been employed to predict investment outcomes based on variables that include funding requests, startup valuations, and team composition, resulting in a 67% prediction accuracy [49]. Logistic regression and neural network models have been used on a dataset of *Shark Tank* deal records, achieving a predictive accuracy of 62.5% in forecasting deal outcomes [50]. Furthermore, tree-based models were used to analyze startup valuations and investor decision-making on *Shark Tank*, from show metadata, with an accuracy of 70% [51].

Beyond classified ML techniques, recent advancements in LLMs have fundamentally transformed the potential for the advanced integration of natural language processing. Today, independent researchers are able to utilize powerful and expensive AI models, which would have taken several years and millions of dollars to develop and deploy in the past. These models particularly excel in the analysis of unstructured textual data, which has traditionally relied on human evaluators.

*4.1 Hypothesis Development*

This research was designed to explore the use of trained AI models (leveraging previous work on the CFA) to provide a solution that addressed the cost, training, and timing constraints contributing to CFA's limited deployment. If successful, this novel approach would allow the deployment of an AI-powered CFA to benefit funds-seeking entrepreneurs and the broader entrepreneurial ecosystem.

The first question to address was whether a trained LLM can replicate the manual CFA assessment in providing valuable insights, previously exclusively done by trained experts. Given the structure of the CFA (i.e., each factor must be independently validated), and the results of our initial analysis, we propose the following hypothesis:

> **H1a:** When evaluating startup pitches, a CFA-augmented AI model would produce consistent and accurate CFA evaluations across each of the eight critical factors, when compared to the scoring of trained human evaluators.

The nature of the eight factors in the CFA is that they are non-compensatory; thus, a positive investment decision is only made if there is no fundamental flaw in any one of the factors, allowing the CFA to predict the actual investment decision on this basis. This allows us to propose the following hypothesis:

> **H1b:** When forecasting an investment decision of a group of BAs, the CFA-augmented AI model will accurately predict the outcome.

There is no doubt that commercially available AI LLM models (such as ChatGPT, Gemini, and Claude) have been effective in helping entrepreneurs develop their ventures and improve every aspect of their business and pitch through the insights they provide. It therefore challenges the usefulness of the CFA-augmented LLM, as it could produce similar results and advice to an unprompted LLM. While it is likely that an LLM is inherently less biased than any randomly selected individual, and can produce fast and cheap feedback, the whole basis of the previous research on the validity of the CFA tool is that this structured approach, analyzing the eight critical factors, provides more accurate decision predictors and advice than an unstructured approach. This leads us to our final set of hypotheses:

> **H2a:** A CFA-prompted LLM would produce better evaluations and investment predictions than an unprompted "stock" LLM (i.e., out-of-the-box GPT-4).
>
> **H2b:** A CFA-prompted LLM would produce better and more relevant feedback and advice to the entrepreneur than an unprompted "stock" LLM.

## 5. Research Methodology

Developing an AI model involves selection of appropriate modeling techniques, data gathering, preparation, model training, and effectiveness validation. We deploy multiple prompt-engineered OpenAI GPT models to assign CFA letter grades to a dataset consisting of transcribed startup pitches from the US TV show *Shark Tank*. The dataset of pitches and their respective CFA letter grades were split into training and testing sets. A series of ML classification models were then trained, tested, and compared across a range of metrics that evaluated their ability to correctly predict funding outcomes (i.e., deal or no deal).

### 5.1 Data collection

This research utilized as its primary data source *Shark Tank*, a US-based television show where entrepreneurs present their business ideas to a panel of typically five investors, referred to as Sharks, keen to make equity funding. The interactions between investors and entrepreneurs are live and unscripted, which reflects real BA decision-making processes influenced by product potential, business models, and financial projections [52, 53]. Maxwell's research addressed the validity of using such data by examining *Dragons' Den* interactions, which are almost identical to *Shark Tank* in format [6]. Maxwell's exposure to the complete recordings of the interactions between the entrepreneurs and the Dragons enabled him to compare the accuracy of the CFA model when evaluating the full live interactions and those edited for TV, noting a very high (95%) "inter-category grader reliability" [54].

A total of 1,153 pitch records were gathered from the publicly accessible *Shark Tank* US Dataset on the AI-data platform Kaggle [55]. This dataset encompassed a variety of industries, pitch styles, and funding results, and comprised 53 identified variables describing the startup. Example variables, or columns (excel), include startup industry, investment amounts, and show season details. Our analysis concentrated on the following key fields: Deal Outcome (binary: 1 for deal secured, 0 for no deal), Ask Amount, and Ask Equity. We omitted the remaining fields because the primary focus was to assign CFA grades based on the verbal pitch, rather than third party-labeled meta data (by Kaggle users). Ask Amount and Equity were retained, as funding opportunities may be rejected based on the inability to reach consensus, which is crucial to funding outcomes. We then obtained subtitle transcripts for 600 pitches from public repositories, including Subdl and OpenSubtitles.org, which were matched to the startup pitches from Kaggle. The data underwent thorough cleaning to rectify missing values, standardize formats, and ensure compatibility with AI modelling. This involved rectifying transcription inaccuracies, eliminating extraneous content, and aligning pitch data with associated deal outcomes.

*5.2 Hypothesis 1a: Synthesizing CFA Features Using AI*

We optimized multiple custom versions of GPT models (GPT 4, 4.1-mini, 4.1, o3) to evaluate each pitch transcript against the CFA criteria by employing advanced techniques, such as agent/workflow orchestration, function calling, context chunking, and prompt engineering. The custom model assigned grades according to the CIC's standardized rubric, with scores varying from A+ to C- (Table 1 and Table 2) [6]. The grades were subsequently codified into structured numerical formats (0 to 10) to facilitate machine learning analysis (Table 2). Maxwell's research indicates that the numbers 3 and 7 were intentionally excluded from the CFA scale to signify that each grade transition denotes a substantial increase in investment readiness, thereby ensuring that scoring differences accurately reflect variations in startup quality [6].

To achieve human-graded CFA scores, we trained three university students to evaluate and grade startup pitches using the CIC's literature and rubric concerning CFA. The trained students then graded the same 31 text transcripts from our database of 600 *Shark Tank* pitches across the eight critical factors. For each pitch, we calculated the arithmetic mean of the three graders' CFA scoring to achieve score consensus.

Table 1. Critical Factor Assessment (CFA) Criteria

| Factor | Criteria | Evaluation Questions |
| --- | --- | --- |
| **Features & Benefits** | Performance Advantages | Does the product offer competitive advantages over current solutions? |

| | Benefits and Costs | Are benefits substantial relative to costs? |
|---|---|---|
| | Customer Demands | Does it meet specific customer needs effectively? |
| **Readiness** | Product Delivery | Is the product ready for market, with key milestones achieved? |
| | Validation Tests | Are there beta tests or customer validations supporting readiness? |
| **Barriers to Entry** | Uniqueness | Does the venture have proprietary tech or strong IP protection? |
| | Market Differentiation | Are there clear competitive advantages? |
| **Adoption** | Customer Engagement | Is there evidence of customer interest or early adoption? |
| | Market Validation | Have customers committed to purchasing? |
| **Supply Chain** | Operational Readiness | Are supply chains and partnerships well established? |
| **Market Size** | Revenue Potential | Is the market size large enough to support high growth? |
| **Entrepreneurial Experience** | Industry Expertise | Does the team have relevant industry or startup experience? |
| **Financial Expectations** | Financial Viability | Are the projections realistic and sustainable? |

Table 2. CFA Grade Conversion Table

| Grade | Score | Adjusted Score |
|---|---|---|
| **A+** | 10 | 80 |
| **A** | 9 | 72 |
| **A-** | 8 | 64 |
| **B+** | 6 | 48 |
| **B** | 5 | 40 |
| **B-** | 4 | 32 |
| **C+** | 2 | 16 |
| **C** | 1 | 8 |
| **C-** | 0 | 0 |
| **N/A** | 0 | 0 |

We analyzed the AI-CFA's ability to evaluate startup pitches by comparing the 31 expert CFA grades (sum of eight factors) to the AI-CFA grades (using four different models: GPT 4, 4.1, 4.1-mini, o3) across three correlation metrics: Spearman's $\rho$ (monotonic rank agreement), Pearson's r (linear correlation), and mean absolute error (average point-wise deviation). This enabled us to determine how accurately a CFA-prompted LLM can replicate human evaluation.

*5.3 Hypothesis 1b: AI Model Assessment of Features*

We employed a multi-model ML pipeline to predict the likelihood of securing funding (binary outcome: deal/no deal) based on the synthesized CFA features. Data preparation began with exploratory data analysis to visualize the dataset and gain insights into existing patterns and characteristics. We discovered a class imbalance, with 66% of pitches resulting in a Yes (deal) investment, which we attempted to address through data sampling techniques.

Next, we generated model features with different CFA factors (i.e., all 8 factors, 8 choose 7, etc.), as well as the requested funding amount and its corresponding equity conversion (Ask $, Ask %) to test the predictive efficacy of different combinations. While Maxwell's research yielded the highest accuracy with all the eight factors [6], it was prudent to conduct an independent evaluation with our model. We chose the following popular, accessible, and efficient classification algorithms for training and evaluation: Logistic Regression, Support Vector Machines (SVM), Naive Bayes Variants, Decision Trees and Random Forests, Gradient Boosting, AdaBoost, XGBoost, CatBoost, Neural Networks, and Linear Discriminant Analysis (LDA). To optimize model performance, we employed hyperparameter optimization within a cross-validation framework, testing different configuration and settings for each model across smaller splits of data, which help with overcoming potential model overfitting. We also tested various ensemble methods, which are common stacking techniques involving the use of multiple base classification models (EG Random Forests and Logistic Regression) in combination. Within the cross-validation optimization, top-performing models were then retrained on the full dataset and tested on a holdout test set (10–20%) to assess generalizability and overall predicative accuracy.

The models were evaluated based on the following performance metrics:
1. Accuracy: Proportion of correctly predicted investment outcomes.
2. F1 score: Balancing precision and recall, especially important for imbalanced datasets.
3. Specificity: Measuring the ability to correctly identify negative cases (i.e., predicting "no deal" accurately).
4. Balanced score: A custom metric averaging F1, accuracy, and specificity for comprehensive evaluation.
5. ROC AUC: The probability that the model will assign a higher "funding likelihood" score to a randomly chosen "deal" pitch than to a randomly chosen "no deal" pitch.

*5.4 Hypothesis 2a/2b: Untrained "Stock" LLM Pitch Processing*

We started a ChatGPT chat using GPT-4 with the following prompt:

"Attached are transcripts from *Shark Tank* pitches, where startups are seeking investments. For each pitch, please analyze and provide a recommendation on whether an investor should consider funding. Include specific reasons for your recommendation based on the pitch details. Additionally, offer targeted advice to each entrepreneur on how they could refine their business model or pitch to enhance their chances of securing an investment. Finally, clearly state a "deal" or "no deal" decision on whether an investor should fund this startup."

The prompt was accompanied with a *Shark Tank* pitch transcript, which was repeated with 31 pitches. We created a new evaluation session for each pitch to ensure that no previous memory or conversations affected the results. The evaluation, recommendation, and deal/no-deal decision were recorded for comparison with our CFA-AI model. The deal/no-deal predictions were compared with the known deal/no-deal outcomes to result in a confusion matrix, comparing the untrained LLM's predictive capabilities. The text evaluation and recommendation were read and compared manually for length of feedback, contextual relevance, depth of insight, and helpfulness.

## 6. Results

### *6.1 Hypothesis 1a/1b: CFA-AI feature extraction and deal prediction*

Across our three-correlation metrics, the CFA-AI factor-grader powered by GPT-4.1-mini, outperformed the rest, and showed a very strong ability to emulate human expert labelling across the eight CFA factors. Table 3 compares the CFA scores (sum of eight) of four GPT model variants against human consensus on a set of 31 pitches.

Table 3. Comparing CFA-AI and Human Expert Grading

| Model | Spearman's ρ | p (Spearman) | Pearson's r | p (Pearson) | Mean Absolute Error (pts) |
|---|---|---|---|---|---|
| **GPT 4.1-mini** | 0.896 | $9.7 \times 10^{-12}$ | 0.886 | $3.4 \times 10^{-11}$ | 4.4 |
| **GPT o3** | 0.840 | $3.4 \times 10^{-9}$ | 0.841 | $3.2 \times 10^{-9}$ | 4.8 |
| **GPT 4** | 0.787 | $1.5 \times 10^{-7}$ | 0.779 | $2.4 \times 10^{-7}$ | 7.9 |
| **GPT 4.1** | 0.780 | $2.3 \times 10^{-7}$ | 0.775 | $3.1 \times 10^{-7}$ | 7.9 |

For monotonic rank agreement (Spearman's ρ), GPT 4.1-mini achieved ρ = 0.896, indicating that its ranking of pitches closely mirrored expert grading (values > 0.8 are considered very strong). Linear correspondence (Pearson's r), r = 0.886, showed that the model's raw score predictions lie close to a best-fit line against the expert scores. For mean absolute error, an average absolute deviation of 4.4 points on an 80-point scale (eight CFA factors x 10 highest possible score) meant GPT 4.1-mini is within ~5 % of the human consensus.

Of the classification algorithms evaluated to test H1b, Soft Voting Ensemble, consisting of Naive Bayes, Logistic Regression, Random Forest, and Gradient Boosting with GPT-4, consistently demonstrated superior performance while utilizing Features & Benefits, Barrier to Entry, Financial Expectations, the total sum of all factors, Ask $ amount, and the respective company equity investors would get. The primary performance metrics for the leading model demonstrated an impressive accuracy of 85.0%, F1 Score of 0.83, Specificity of 0.80, ROC AUC of 0.82, Precision of 0.83, and recall of 0.84 (Table 4, Figure 1 for confusion matrix). Additional models also showed noteworthy performance, as seen in Table 4.

Table 4. Performance Metrics for Top Models

| Model | Feature Combination | Accuracy (%) | F1 Score | Specificity | ROC AUC | Precision | Recall |
|---|---|---|---|---|---|---|---|
| **Soft Voting Ensemble – Naive Bayes, Logistic Regression, Random Forest, Gradient Boosting (GPT-4)** | 1, 3, 8, Total, Ask $, Ask % | 85.0 | 0.83 | 0.80 | 0.82 | 0.83 | 0.84 |
| **CatBoost (GPT-4.1-mini)** | 1, 4, Total, Ask $, Ask % | 85.0 | 0.82 | 0.70 | 0.76 | 0.84 | 0.81 |
| **CatBoost (GPT-4.1-mini)** | All 8, Total, Ask $, Ask % | 78.0 | 0.75 | 0.6 | 0.74 | 0.76 | 0.74 |
| **Soft Voting Ensemble – Same as above (GPT-4)** | All 8, Total, Ask $, Ask % | 75.0 | 0.72 | 0.65 | 0.76 | 0.72 | 0.73 |

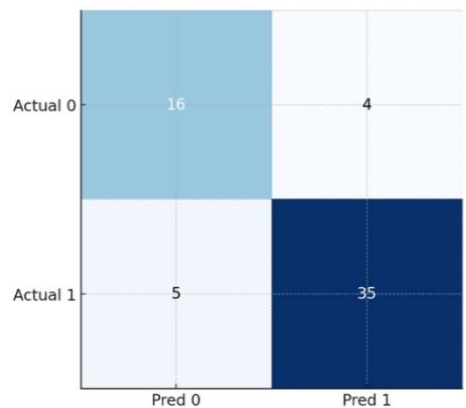

Figure 1. Soft Voting Ensemble – Naive Bayes, Logistic Regression, Random Forest, Gradient Boosting (GPT-4) (1, 3, 8, Total, Ask $, Ask %) Confusion Matrix

A feature importance analysis on our top performing stacking ensemble model showed that the total CFA score (importance weight = 0.195) was the single strongest predictor, followed by Financial Expectations (0.158), Ask Amount (0.148), Features & Benefits (0.145), Barrier to Entry (0.101), and Ask Equity (0.077). The results demonstrated that the model effectively identified both successful (deal) and unsuccessful (no deal) pitches with 85.0% accuracy, exhibiting strong performance across all key ML metrics, and supporting H1a/H1b directly. If the LLM model was not effectively assigning CFA grades, the overall predictive accuracy would have suffered.

### *6.2 Hypothesis 2a/2b: CFA-Prompted LLM vs. Unprompted LLM*

While an untrained GPT-4 showed some predictive capability (accuracy of 58.1%), it was outperformed by our CFA-prompted LLM on the same test dataset (N=31, Accuracy = 77.4%) (Table 6 and Figure 2). The largest predictive gap was seen in the Specificity (of all *Shark Tank* "no deal" pitches, the percentage our model correctly labeled as "no deal") of the two models (0.39 vs. 0.92). Furthermore, the CFA-LLM generated contextual evaluations and recommendations in full sentences and paragraphs, addressing various aspects of the pitch in detail, drawing specific examples from the pitch and generating relevant and sophisticated recommendations. In comparison, the GPT-4-generated evaluations and recommendations were presented in short bullet-point format, and were not consistent across various reports, often omitting crucial feedback or insights. The results therefore support H2a/2b.

Table 6. CFA-Prompted LLM vs. Unprompted LLM Performance Metrics (N=31)

| Metric | GPT-4 Unprompted | CFA-Prompted (GPT-4) |
|---|---|---|
| Accuracy | 58.1% | 77.4% |
| Precision (Deal) | 0.62 | 0.92 |
| Recall (Deal) | 0.72 | 0.67 |
| F1 Score | 0.55 | 0.77 |
| Specificity (No Deal) | 0.39 | 0.92 |

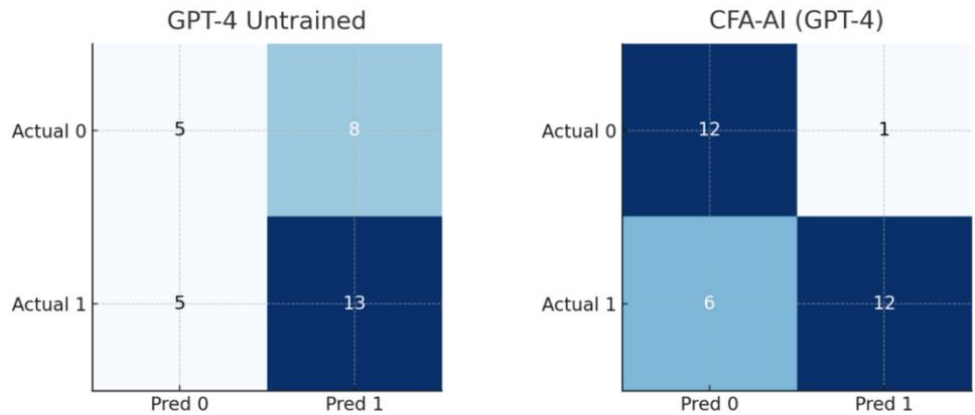

Figure 2. CFA-Prompted LLM vs. Unprompted LLM Confusion Matrices (N=31)

*6.3 Error Analysis*

For the 31 pitches and 8 factors evaluated by the CFA-prompted GPT4 model, Factors 1, 3, 5 exhibited the largest mean difference when comparing to human evaluations (Table 8). For the same 31 pitches, Table 9 displays the seven where CFA-AI misclassified the final deal outcomes (FN – False Negative, FP – False Positive).

Table 8. Factors with Largest CFA Score Differences, CFA-prompted GPT4 vs. Humans

| Factor | Mean difference (pts) | Pearson *r* |
|---|---|---|
| **F5 – Supply Chain** | +3.2 | 0.49 |
| **F3 – Barrier to Entry** | +2.5 | 0.43 |
| **F1 – Features & Benefits** | +1.6 | 0.38 |

Table 9. Misclassified Deal Outcomes of 31 Pitches by CFA-Prompted GPT4

| Pitch # | Outcome | AI Confidence (> 50%=Deal) | GPT-4 total | Human total | Error type |
|---|---|---|---|---|---|
| **3** | 1 | 35.8 % | 57 | 57 | FN |
| **12** | 1 | 17.1 % | 27 | 41 | FN |
| **14** | 1 | 43.2 % | 51 | 50 | FN |
| **17** | 1 | 48.3 % | 44 | 42 | FN |
| **27** | 1 | 20.5 % | 36 | 40 | FN |
| **29** | 1 | 18.3 % | 37 | 45 | FN |
| **10** | 0 | 50.5 % | 61 | 51 | FP |

## 7. Discussion

### 7.1 Comparison with Andrew Maxwell's CFA

The use of AI in combination with the CFA allowed us to build on Maxwell's research on BA decision making and propose a novel, more efficient, scalable, and accessible approach to assessing early-stage ventures. To evaluate AI's validity to deploy Maxwell's CFA, we first determined whether it could replicate human-expert results. Maxwell's CFA method, dependent on human evaluators, demonstrated notable predictive performance, attaining an overall accuracy of 87.5% in forecasting early-stage investment decisions [6]. Our integrated CFA-AI model demonstrated a test accuracy of up to 85% when employing ensemble voting models to predict BA funding outcomes. However, it is important to highlight several differences in data sources and evaluation methods between the two. Maxwell's study reviewed extensive conversations between entrepreneurs and investors, lasting 30–60 min, offering evaluators more context and potential signals for assessment [6]. By contrast, our model utilized brief 3–5-min *Shark Tank* pitches, which offer significant reliability (as discussed earlier in the research) but could still lack depth. Maxwell's model combined evaluations from several trained CFA raters, adding a dimension of human consensus that likely enhanced its accuracy [6]. This can be mimicked in our model through the future integration of multi-agent models. Our CFA-AI model utilized a single output generated by an LLM for each pitch, which may introduce variance stemming from model-specific biases.

Maxwell's model demonstrated efficacy in predicting negative decision outcomes, achieving 100% accuracy in identifying pitches that were likely to be rejected based on fatal flaws (compared to our model's 80% specificity) [6]. The CFA-AI model demonstrated enhanced efficacy in forecasting favorable results (Yes Deal). This divergence may suggest variations in the data context. Pitches aired on TV often highlight success-oriented narratives designed to attract investor interest and audience engagement. Maxwell's CFA study employed a dataset of pitches that were not solely broadcasted, thus providing a more precise depiction of the deal/no-deal distribution.

### 7.2 Comparison to Other Shark Tank Studies

Our model's performance surpassed that of previous studies utilizing *Shark Tank* data, with predictive accuracies ranging between 62.5–70%:

- Deodhar et al. utilized ANNs and random forests on a dataset of 117 pitches from *Shark Tank India*, achieving an accuracy of 67% [49].
- Sherk et al. applied logistic regression and neural networks to *Shark Tank* data, achieving a predictive accuracy of 62.5% [50].
- Lavanchy et al. employed decision tree models to analyze startup valuations and investor decision-making, attaining an accuracy of 70% [51].

The improved performance of our model can be attributed to two main factors. First, while prior research frequently relied on structured metadata, such as deal size, team demographics, and industry type, our methodology utilized LLM-based feature extraction to examine the entirety of the textual content in startup pitches. This method captures nuanced and context-dependent information that structured metadata may overlook. Second, by integrating the CFA framework as a systematic evaluative tool, our model focuses on empirically validated, domain-specific factors instead of generic indicators. The integration of comprehensive, text-based features with the structured CFA framework represents a notable methodological improvement, enhancing our work's standing relative to previous investment evaluation methods.

### 7.3 Analyzing the Results

*7.3.1 Top-Performing Model*

The top-performing model, combining the LLM-based CFA factor synthesis with soft voting ensemble, exhibited Recall and F1 scores of 0.84 and 0.83 respectively. Recall accounts for the total pitches that received an investment, compared to how many our model identifies correctly (TP/(TP+FN)). The F1 score denotes a harmonic mean between precision (of pitches that our model thought will receive investment, how many did) and recall. Our model's specificity (TN/(TN+FP)) of 0.80 denoted its ability to correctly identify a significant portion of non-deal pitches. The overall accuracy of 0.85 (85%) denotes the total number of correct predictions divided by the total number of pitches in test set. In a test set of 60, the model correctly classified 51 pitch outcomes. These results indicate that the model did not blindly classify all pitches as "deal" cases, often a sign of an overfitted model, but effectively differentiated between investment outcomes.

A potential explanation for the strong performance of Naive Bayes (as a top-performing base model in our ensemble) is its robustness with smaller datasets, and the features derived from text. The probabilistic nature enables effective management of class imbalance, and the assumption of feature independence corresponds with the structure of a CFA-based evaluation, wherein each factor is evaluated individually prior to aggregation.

*7.3.2 Importance of Specific CFA Factors*

An analysis of features among the two top-performing models identified Features & Benefits (1), Barrier to Entry (3), Adoption (4), and Financial Expectations (8) as the most predictive individual factors, alongside Total CFA Score, Ask Amount, and the respective Ask Percent (e.g. $100,000 in exchange for 10% of the company). Readiness (2), Supply Chain (5), Market Size (6), and Entrepreneurial Experience (7) were missing, which may be explained either by our model's shortcomings, or by the nature of the information typically expressed in the transcript of the *Shark Tank* interaction. Startups that make it to the show have already exhibited some level of Readiness and thereby Supply Chain. Most of the startups displayed strong product readiness and some level of

sales, which implies that the product is in the later stages of readiness and basic supply chain processes were in place.

While Maxwell's CFA deemed eight factors important, it is possible that the show producers chose to omit specific discussion points from aired pitches. Certain factors may also have been pre-screened prior to selection of pitches that will air, with Entrepreneur Experience a likely candidate, to ensure the presenter is engaging and relevant. Such factors would limit our model's ability to derive crucial decision-making data, which explains the omission of certain factors from top performing models. Finally, the training and prompting of our LLM-based feature extraction may have been inadequate for underperforming factors, indicating the need for further finetuning and prompt engineering.

### 7.3.3 Comparing Performance of Different GPT Models

It is logical to assume that new and more advanced LLMs will outperform incumbents (e.g., GPT 4.1 vs. GPT 4), as regularly shown in popular LLM benchmarking reports such as LiveBench [56]. While GPT4.1-mini outperformed GPT4 in its ability to replicate human grading (Spearman's $\rho$ 0.896 vs. 0.787), it did not replicate its superiority in predicting pitch funding outcomes. GPT4 outperformed in two key metrics — Specificity and ROC AUC (the model's ability to give a higher funding likelihood to a *Shark Tank*-funded pitch when comparing one deal vs. one no-deal pitch at random), by 10 and 6 percentage points respectively. The result gap yielded two potential hypotheses: poor quality of human grading, or the failure to find an optimal model configuration for the CFA factors extracted by GPT4.1-mini. While we assert that newer AI models outperformed across common benchmarks, such as coding and reasoning, there is no evidence that these improvements would apply to all applications and domain — text context classification using the CFA rubric being one of them.

### 7.3.4 CFA-Prompted LLM Model vs. Unprompted GPT-4

Our results demonstrated the superior performance and usefulness of a custom-prompted LLM with a research-backed decision-making framework. During initial phases of our research, a common criticism was: "Why is your tool better than simply using ChatGPT?" The answer was clear while comparing it to CFA-AI with 31 pitches (77.4% Accuracy for CFA-AI vs. 58.1% for an untrained GPT-4). While an untrained GPT-4 may excel in a broader context of idea and text generation, it may also suffer from a lack of focused training in identifying investment opportunities correctly, and providing optimally useful pitch feedback and improvement recommendations.

### 7.5 Research Limitations

#### 7.5.1 Dataset Limitations

The dependence on *Shark Tank* data, although a comprehensive and accessible dataset, presented possible biases. The program is structured for entertainment, potentially affecting the manner in which entrepreneurs deliver their pitches and investor reactions, as discussed in Subsection 7.1. This constraint may have influenced the model's capacity to capture nuanced decision-making factors, especially those associated with team dynamics or investor follow-up inquiries. Furthermore, the dataset of pitches and their respective outcomes did not take into account additional variables that contribute to BA decision making, such as any disagreement about the startup's valuation or dishonesty. During the review of our model's results, we noticed a case of deal rejection owing to an inflated valuation and investment ask from an entrepreneur, which was not reflected in the initial evaluated pitch and was not covered by a CFA factor.

*7.5.2 Class Imbalance*

Class imbalance refers to the unequal distribution of instances across different categories in a dataset, which can adversely affect the performance of ML models. The dataset demonstrated a class imbalance, with 66% of pitches leading to deals and 34% resulting in none. Despite the application of techniques such as random undersampling to address this issue, class imbalance may have continued to affect the model's generalization capabilities, especially in predicting negative outcomes.

## 8. Implications, Practical Applications, and Future Research

*8.1 CFA Factor Synthesis Using OpenAI Competitors*

A key contribution of this research is validating using an LLM to extract CFA features from startup pitches. While publicly available LLMs such as GPT-4 are designed with the goal of facilitating broad natural language generation, our research demonstrated that a prompted model could execute complex and specific evaluative tasks. This was made possible through prompt engineering using the CFA rubric, as well as the intelligent architecture of OpenAI API calls, using varying context windows, model temperatures, and so on. Utilizing competitor models, such as Anthropic's Claude or Google's Gemini, may yield different results, as they are built using differing architectures and trained with different datasets. Comparing our GPT-4 and GPT-4.1-mini to additional OpenAI models (o3, o1, etc.) would offer new data and results.

*8.2 Application for Entrepreneurs and Educators*

While not evaluated in-depth in the Section 6 (Results), our CFA-AI model (publicly available under Expitch.com) provided text evaluation based on the criteria for each of the eight factors, as well as text recommendations for attaining an A+ rating in each critical factor. This resource functions as an educational tool for entrepreneurship courses, which we have validated through dozens of conversations with academics, incubators, and accelerators. Expitch has been used in multiple hackathons by hundreds of students. Future research can examine the usefulness and effectiveness of the CFA in facilitating conversations with startup mentors, by way of providing actionable metrics to track and work on over time.

Future research may include longitudinal studies to evaluate the effects of AI-generated feedback on entrepreneurial outcomes. Entrepreneurs could be monitored longitudinally to assess improvements in investment readiness and success rates after applying AI-driven recommendations. Controlled experiments comparing groups with and without AI feedback could yield causal evidence regarding the tool's effectiveness. Complementary qualitative research such as surveys can investigate entrepreneurs' perceptions on the tool's helpfulness.

### *8.3 Application for Investors, Policymakers, and Ecosystem Builders*

The CFA-AI model provides an efficient and uniform approach for the initial assessment of investment prospects. Although seasoned investors often depend on their intuition and judgment for final decisions, the model can function as an efficient triage tool for angel groups, venture capital firms, and accelerators. This approach aids in minimizing cognitive biases, optimizing due diligence processes, and ensuring that valuable opportunities are not missed due to human error or information overload. Although angel investors provided valuable inputs during the development of our model, future research could investigate the practical benefits of deploying the CFA-AI tool in real-world investment settings. Rather than solely predicting startup success or failure, this tool has the potential to function as an advanced decision-support system, enabling BAs to efficiently identify promising investment opportunities. Given that human judgment is frequently compromised by cognitive biases such as overconfidence, anchoring, and confirmation bias [7], the integration of AI assessments may serve to mitigate these limitations.

To rigorously test this proposition, future studies could employ experimental designs in which angel investors make investment decisions both with and without the assistance of the CFA-AI tool. Additionally, researchers could investigate the impact on deploying AI decision-making tools to mitigate intrinsic human bias and subconscious decision-making factors. The insights provided can guide policy development in entrepreneurial finance, especially in creating programs to support early-stage ventures. The AI model may be utilized in public funding initiatives, startup competitions, or grant evaluations to facilitate objective, data-driven decision-making.

### *8.4 Refinement of Feature Extraction Techniques*

Future research should explore alternative advanced NLP methods to further improve feature extraction. Sentiment analysis, currently not covered by our GPT-4-based model, can be employed to assess the emotional tone and confidence levels in startup pitches. Techniques such as topic modelling can identify prevailing themes and subjects in pitches, assessing their alignment with current market trends and investor interests. While we lightly experimented with topic modeling and sentiment analysis, our dataset of 600 pitches was insufficient to capture text patterns.

By modifying the API adaptation to our tool, future research can investigate the effect of alternative LLMs on model performance. LLMs such as Google's Gemini and Anthropic's Claude offer unique architectural advantages, such as better contextual comprehension and improved finetuning abilities. More advanced methodologies may include agent-orchestration of multiple different LLMs to discuss CFA grades, similar to how the BAs on *Shark Tank* or *Dragons' Den* interact. OpenAI is offering models capable of conducting online research, which may be useful in reinforcing and informing the assignment of CFA grades based on researched data.

### *8.5 Ethical Considerations and Bias Mitigation*

As AI models increasingly influence critical decisions in entrepreneurial finance, future research must thoroughly examine the ethical challenges involved. It is essential to examine algorithmic bias, particularly to assess whether AI models unintentionally introduce or perpetuate biases associated with gender, race, or socioeconomic status in the evaluation of startups. Enhancing transparency and explainability is essential, and both entrepreneurs and investors must be able to interpret AI-generated assessments to foster trust and accountability. Establishing robust data privacy protocols to protect sensitive business information, particularly as AI tools are integrated into real-world investment processes, is crucial. Systematic bias audits and the development of comprehensive frameworks for explainable AI can mitigate risks, ensuring that the CFA-AI model is both effective and ethically responsible.

### *8.6 Practical Spillovers*

The availability of AI tools to train and educate entrepreneurs will transform how we support entrepreneurship, providing real time on-demand tools that:

- Lower the cost significantly, while improving the speed of feedback. Based on Maxwell's original assessment, the original CFA cost was 3 raters × 2 days or about $1400 per venture; while the AI version costs less than US$1 and is completed in seconds.
- Allow real-time iteration providing scores and direct feedback/diagnostics for founders to iteratively learn each component of the CFA. Expitch is already being used in classroom pilots and at industry pitch events.
- Enhance the use of toolbox and frameworks that have been developed by researchers, enabling a similar prompt-engineering templates to be used to bring other under-used rubrics (e.g., TRL, JTBD, BMC) into daily practice.